\documentclass[pdflatex,sn-mathphys-num]{sn-jnl}

\usepackage[utf8]{inputenc}

\usepackage{graphicx}%
\usepackage{multirow}%
\usepackage{amsmath,amssymb,amsfonts}%
\usepackage{amsthm}%
\usepackage{mathrsfs}%
\usepackage[title]{appendix}%
\usepackage{xcolor}%
\usepackage{textcomp}%
\usepackage{manyfoot}%
\usepackage{caption}
\usepackage{booktabs}%
\usepackage{algorithm}%
\usepackage{algorithmicx}%
\usepackage{algpseudocode}%
\usepackage{listings}%
\usepackage{makecell}
\usepackage{booktabs,tabularx,siunitx,xcolor}
\usepackage[table]{xcolor}%
\definecolor{good}{HTML}{E6F4EA} 
\definecolor{bad}{HTML}{FDECEA}  
\usepackage[dvipsnames]{xcolor}
\definecolor{VideoPink}{HTML}{EA048C}
\newcommand{\good}[1]{\cellcolor{good}\bfseries #1}
\newcommand{\bad}[1]{\cellcolor{bad}\bfseries #1}

\theoremstyle{thmstyleone}%
\theoremstyle{thmstyletwo}%

\theoremstyle{thmstylethree}%

\begin{document}

\title[ARport: An Augmented Reality System for Markerless Image-Guided Port Placement in Robotic Surgery]{ARport: An Augmented Reality System for Markerless Image-Guided Port Placement in Robotic Surgery}


\author[1]{\fnm{Zheng} \sur{Han}}\email{zheng-han@link.cuhk.edu.hk}
\equalcont{These authors contributed equally to this work.}

\author[1]{\fnm{Zixin} \sur{Yang}}\email{zixinyang@mrc-cuhk.com}
\equalcont{These authors contributed equally to this work.}

\author[1]{\fnm{Yonghao} \sur{Long}}\email{yhlong@cse.cuhk.edu.hk}

\author[2]{\fnm{Lin} \sur{Zhang}}\email{jim.zhang@csrbtx.com}

\author[3]{\fnm{Peter} \sur{Kazanzides}}\email{pkaz@jhu.edu}

\author*[1]{\fnm{Qi} \sur{Dou}}\email{qidou@cuhk.edu.hk}

\affil*[1]{\orgdiv{Department of
Computer Science and Engineering}, \orgname{The Chinese University of Hong
Kong}, \orgaddress{\city{HKSAR}, \postcode{999077}, \country{China}}}

\affil[2]{\orgname{Cornerstone Robotics Ltd.}, \orgaddress{\city{HKSAR}, \postcode{999077}, \country{China}}}

\affil[3]{\orgdiv{Department of Computer Science}, \orgname{Johns Hopkins
University}, \orgaddress{\city{Baltimore}, \state{MD}, \postcode{21218},  \country{USA}}}


\abstract{\textbf{Purpose:} Precise port placement is a critical step in robot-assisted surgery, where port configuration influences both visual access to the operative field and instrument maneuverability. To bridge the gap between preoperative planning and intraoperative execution, we present ARport, an augmented reality (AR) system that automatically maps pre-planned trocar layouts onto the patient’s body surface, providing intuitive spatial guidance during surgical preparation.

\textbf{Methods:} ARport, implemented on an optical see-through head-mounted display (OST-HMD), operates without any external sensors or markers, simplifying setup and enhancing workflow integration. It reconstructs the operative scene from RGB, depth, and pose data captured by the OST-HMD, extracts the patient’s body surface using a foundation model, and performs surface-based markerless registration to align preoperative anatomical models to the extracted patient’s body surface, enabling \textit{in-situ} visualization of planned trocar layouts. A \href{https://zhenghan98.github.io/video_demo/ARport/demo.mp4}{\textcolor{VideoPink}{demonstration video}} illustrating the overall workflow is available online.

\textbf{Results:} In full-scale human-phantom experiments, ARport accurately overlaid pre-planned trocar sites onto the physical phantom, achieving consistent spatial correspondence between virtual plans and real anatomy.

\textbf{Conclusion:} ARport provides a fully marker-free and hardware-minimal solution for visualizing preoperative trocar plans directly on the patient’s body surface. The system facilitates efficient intraoperative setup and demonstrates potential for seamless integration into routine clinical workflows.
}


\maketitle

\section{Introduction}\label{Introduction}

\begin{wrapfigure}{r}{0.45\textwidth}
    \centering
    \vspace{-20pt} 
    \includegraphics[width=0.45\textwidth]{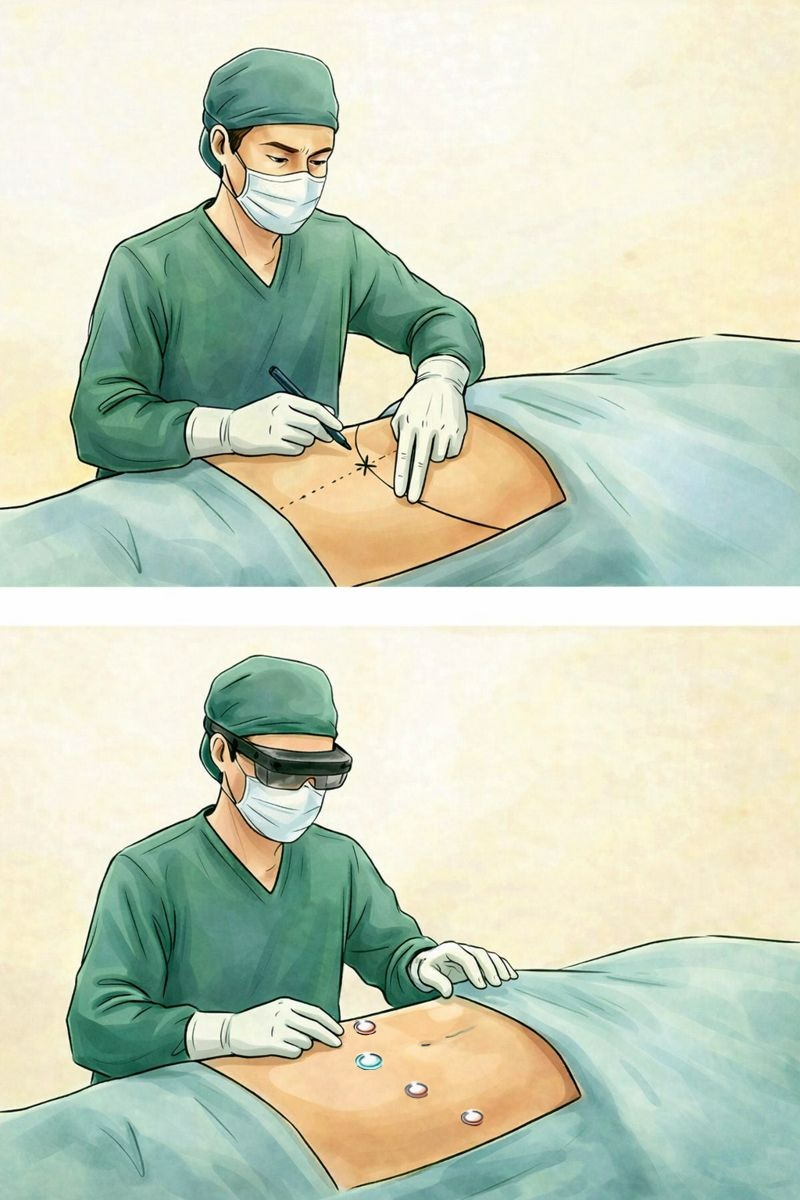}
    \caption{Port placement in robotic surgery. Upper: Heuristic, experience-dependent port placement. Bottom: AR-assisted port placement with intuitive \textit{in-situ} guidance.}
    \label{fig:teaser}
    \vspace{-20pt}
\end{wrapfigure}

Port placement is a critical step in minimally invasive robot-assisted surgery, as it directly affects instrument reachability, surgeon dexterity, and overall procedural efficiency~\cite{coste2004optimal}.
Although optimal port layouts are typically planned preoperatively using CT or MRI data, surgeons currently lack standardized intraoperative guidance to accurately translate these plans onto the patient’s body surface~\cite{hayashi2017optimal}.
As a result, port sites are often determined heuristically based on the surgeon’s experience and anatomical landmarks, making port placement an experience-dependent process (Fig.~\ref{fig:teaser}, upper). This practice can lead to suboptimal port configurations that restrict instrument maneuverability or necessitate additional intraoperative ports, thereby increasing surgical complexity and patient trauma. Therefore, a system that provides intuitive and reliable intraoperative guidance for port placement is critically needed.

An effective port placement system should be easy to set up and capable of providing intuitive 3D context and real-time visual feedback in the operating room. 
However, existing systems remain limited. 
Early approaches primarily relied on fixed display screens~\cite{hayashi2017optimal} or 2D laser projections~\cite{simoes2013leonardo}, which failed to deliver immersive 3D perception or surgeon-centered interaction. 
Moreover, some methods required complex calibration or additional tracking hardware~\cite{feuerstein2008intraoperative}, hindering their adoption in routine clinical workflows. 
For instance, Feuerstein \textit{et al.}~\cite{feuerstein2008intraoperative} introduced a comprehensive port placement framework integrating fiducial markers, an optically tracked laparoscope, and an intraoperative C-arm for image-based localization; yet, its reliance on extensive hardware and laborious setup limited its practicality. 
Laser-based systems such as Leonardo~\cite{simoes2013leonardo} projected port plans onto the patient’s abdomen but could not convey true 3D depth and required uniform projection surfaces, further restricting usability.

In comparison, optical see-through head-mounted displays (OST-HMDs) offer a more streamlined solution by integrating \textit{in-situ} visualization with surgeon-centered spatial awareness. Recently, Agnino \textit{et al.} \cite{agnino2025clinical} proposed an intraoperative port localization system utilizing the HoloLens 2, which optimized port positioning with enhanced visualization; however, the system relied on manual alignment, which disrupted the procedural flow and introduced uncertainty in the accuracy of port site localization.

\textbf{Contributions.} To address these gaps, we present ARport, an artificial intelligence (AI)-powered augmented reality (AR) system designed for intuitive and automatic port localization. Our main contributions are summarized as follows:

\begin{enumerate}
  \item \textbf{An intuitive AR guidance system for port placement.} 
  We develop ARport, an AR-based port placement guidance system that superimposes preoperative trocar plans onto the patient’s body surface, providing direct \textit{in-situ} visual guidance during the surgical setup phase (Fig.~\ref{fig:teaser}, bottom).

  \item \textbf{An AI-enabled, and markerless implementation.}
  We propose an intelligent AR framework that integrates AI perception with the OST-HMD. This approach achieves AR registration without relying on external sensors or markers, thereby minimizing setup complexity and enabling seamless clinical integration.

  \item \textbf{Validation on a full-scale surgical phantom.} 
  We validate the complete ARport workflow on a human-scale surgical phantom under realistic conditions, demonstrating both its feasibility and accuracy.
\end{enumerate}

\section{Method}\label{Method}

\begin{figure*}[th]
\centering
\includegraphics[width=1.0\textwidth]{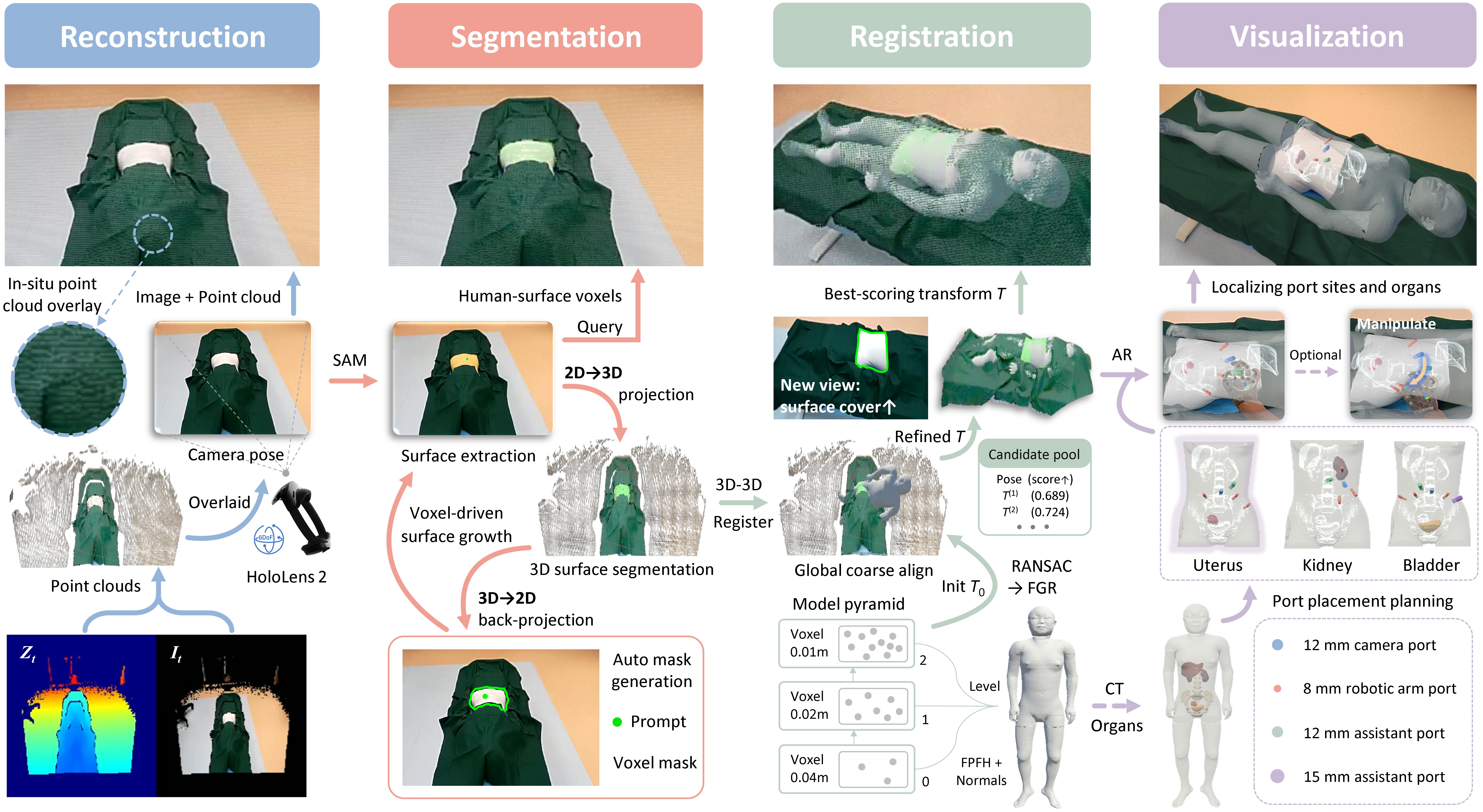}
\caption{The ARport system integrates reconstruction, segmentation, registration, and visualization on the HL2. It reconstructs the operative scene from raw RGB–depth–pose streams, extracts the abdominal surface through a foundation model, performs surface-based registration, and visualizes port sites and organs \textit{in-situ} to enable AR-guided localization and adjustment of trocar positions. A \href{https://zhenghan98.github.io/video_demo/ARport/demo.mp4}{\textcolor{VideoPink}{video demo}} showcasing the overall workflow is available online.}
\label{fig:framework}
\end{figure*}

\noindent We developed an AR system on the Microsoft HoloLens~2 (HL2) to enable intuitive port placement by visualizing preoperative trocar plans directly on the patient, as illustrated in Fig.~\ref{fig:framework}. The system integrates reconstruction, segmentation, registration, and visualization in a sequential pipeline. HL2 streams RGB, depth, and six-degree-of-freedom (6-DoF) pose data, which are fused into a colored point cloud for real-time 3D reconstruction of the scene (see \S\ref{Real-world 3D Mapping}). Based on this reconstruction, the human surface is extracted by initializing a 3D voxel mask from a 2D segmentation based on the Segment Anything Model (SAM)~\cite{kirillov2023segment} and dynamically refining it through an iterative 3D-2D feedback loop (see \S\ref{sec:human-surface}). The extracted surface then guides CT-to-scene registration, aligning the preoperative 3D mesh with the reconstructed scene. Finally, the registered mesh is rendered \textit{in-situ} through AR visualization, overlaying pre-planned port sites and internal organs onto the patient’s body (see \S\ref{subsec:registration}).

\subsection{Real-world 3D Reconstruction}
\label{Real-world 3D Mapping}

This module reconstructs a colored 3D map of the environment in a consistent world coordinate system, providing a spatial foundation for subsequent segmentation and registration, as illustrated in the \textit{Reconstruction} panel of Fig.~\ref{fig:framework}.
At each time step \(t\), the HL2 streams a photo/video (PV) image \(I_t\), a long-throw (LT) depth map \(Z_t\), and the reference-to-world pose \(T_{\mathrm{ref}\to w}(t)\in SE(3)\), where the reference frame denotes the HL2 device coordinate system shared by all onboard sensors. 

Each depth pixel \(\mathbf{u}=(u,v)\) with depth value \(z=Z_t(u,v)\) is back-projected using the LT camera intrinsics \(K_d\) and transformed into the world frame as:

\begin{equation}
\mathbf{x} = T_{\mathrm{ref}\to w}(t)\,T_{d\to \mathrm{ref}}\,\pi^{-1}(K_d,\,\mathbf{u},\,z),
\label{eq:backproject}
\end{equation}

\noindent where each 3D point \(\mathbf{x}\) is then projected into the PV image to retrieve its color \(\mathsf{c}=I_t(\tilde{\mathbf{u}})\), forming the colored point cloud
\(
X_t = \{(\mathbf{x}_i, \mathsf{c}_i)\}_i
\) for better visualization. The world-aligned point cloud is voxel down-sampled with resolution \(\delta_{\mathrm{send}}\) for efficient streaming, and the pair \((X_t,\,T_{\mathrm{ref}\to w}(t))\) is forwarded to the subsequent modules. 

\begin{figure*}[h]
    \centering
    \includegraphics[width=1.0\textwidth]{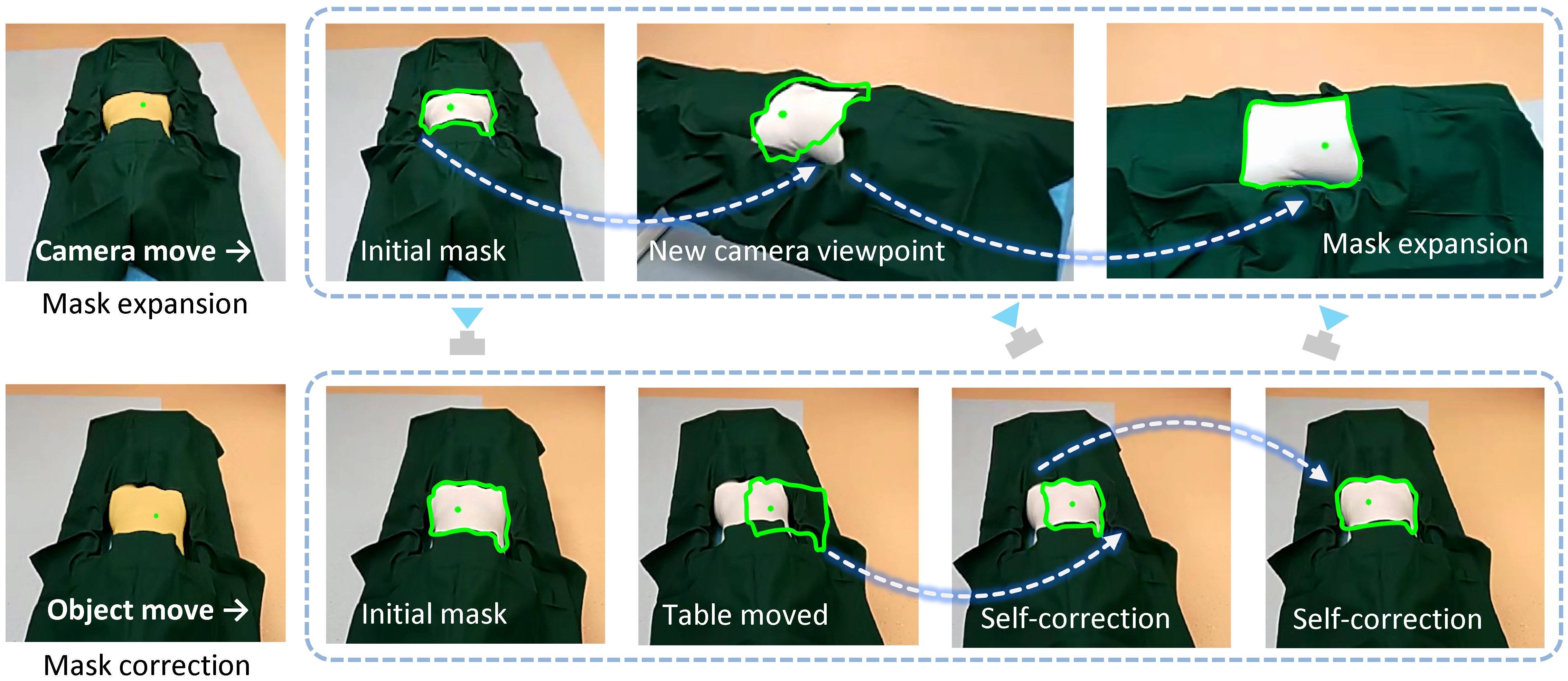}
    \caption{Illustration of segmentation mask expansion and self-correction. 
    Upper: new viewpoints enable progressive mask growth. 
    Bottom: motion or occlusion triggers automatic self-correction.
   A \href{https://zhenghan98.github.io/video_demo/ARport/mask.mp4}{\textcolor{VideoPink}{video demonstration}} of the mask expansion and self-correction mechanism is available online.}
    \label{fig:MASK}
\end{figure*}

\subsection{Human Surface Segmentation and Tracking}
\label{sec:human-surface}

\noindent Following the real-world 3D mapping described in Sec.~\ref{Real-world 3D Mapping}, this module extracts and maintains the mask of the target human surface from the reconstructed scene. 
The mask is dynamically tracked as the camera viewpoint changes and corrected when the object moves, as illustrated in Fig.~\ref{fig:MASK}. 
A 3D voxel mask is maintained in real time and updated through three consecutive stages: 
(1) \textbf{Mask initialization}, which generates a 3D surface mask from a single 2D segmentation; 
(2) \textbf{Mask propagation}, which refines and tracks the mask through iterative 3D-2D feedback; and 
(3) \textbf{Mask regularization}, which ensures geometric consistency and removes erroneous regions.

\textbf{Mask initialization (2D$\rightarrow$3D projection).}
The pipeline begins by initializing the 3D mask from a single 2D segmentation. 
A few user clicks on the PV frame trigger a single SAM~\cite{kirillov2023segment} inference, producing a binary mask \(M_t^{\mathrm{init}}\). 
Each point \(\mathbf{x}\in X_t\) in the reconstructed point cloud is projected into the PV image, and those whose projections fall within \(M_t^{\mathrm{init}}\) are retained:
\begin{equation}
X_t^{\mathrm{init}} = \{\mathbf{x}\in X_t \mid M_t^{\mathrm{init}}(\pi(K_{\mathrm{pv}}(t),T_{\mathrm{pv}\to w}(t),\mathbf{x}))=1\}.
\end{equation}
The resulting 3D subset is voxelized to form an initial mask \(V_r\), which seeds the subsequent online refinement process as new sensor data arrives.

\textbf{Mask propagation (Iterative 3D-2D feedback).}
Once initialized, the voxel mask is dynamically refined as the HL2 streams new frames. 
Each update leverages a feedback loop between 3D geometry and 2D appearance cues to maintain temporal consistency.
Specifically, active voxel centers are projected onto the current PV frame to generate a visibility mask $M^{\mathrm{proj}}_{r,t}$, from which a high-confidence point is selected to prompt SAM for a new segmentation. 
The resulting segmentation is aligned with depth gradients, and valid pixels from the refined mask $\widehat{M}_{r,t}$ are back-projected to the active frontier $\partial V_r$. 
New voxels are instantiated only after repeated observations, while existing ones update their occupancy statistics to reflect tracking confidence.

\textbf{Mask regularization.}
As mask propagation continues, accumulated noise or false expansion may occur, motivating a regularization step to prune inconsistent regions. First, each new segmentation is validated to ensure geometric consistency. 
An update is accepted only if the inferred mask $\widehat{M}_{r,t}$ maintains sufficient overlap with the projected current surface silhouette:
\begin{equation}
\mathrm{IoU}(\widehat{M}_{r,t}, M^{\mathrm{proj}}_{r,t}) \ge \tau_{\mathrm{IoU}},
\end{equation}
where \(\tau_{\mathrm{IoU}}\) denotes the minimum allowed intersection-over-union threshold.  

Further, erroneous voxels are pruned based on probabilistic occupancy. Voxels with conflicting observations are gradually carved out:
\begin{equation}
\frac{H_{\mathrm{free}}(v)}{H_{\mathrm{occ}}(v)+1} > \rho_{\max},
\end{equation}
where $H_{\mathrm{free}}$ and $H_{\mathrm{occ}}$ track the number of times a voxel is observed as free or occupied, respectively, and $\rho_{\max}$ serves as the pruning sensitivity threshold. 

\subsection{CT-to-Scene Registration}
\label{subsec:registration}

\noindent Following human surface extraction (Sec.~\ref{sec:human-surface}), this module estimates the rigid transformation $T = [\,R \mid \mathbf{t}\,] \in SE(3)$  aligning the preoperative CT mesh $M$ with the intraoperative scene points $O_t$. 
Both point sets are processed into multi-resolution pyramids $\{M_\ell, O_{t,\ell}\}$ to facilitate a coarse-to-fine alignment strategy.

\textbf{Pose initialization.}
We first estimate a global alignment $T_0$ at the coarsest level using fast point feature histograms (FPFH)~\cite{rusu2009fast} via RANSAC ICP, falling back to Fast Global Registration (FGR)~\cite{zhou2016fast} if the RANSAC step fails.
To resolve anatomical symmetries common in human bodies, we maintain a candidate pool $\mathcal{T}$ comprising the temporal tracking prior, the feature-based initializer $T_0$, and their $\pi$-rotations about canonical and local axes. 
This diverse hypothesis set ensures recovery from flipped or mirrored initialization states.

\textbf{Multi-scale refinement.}
Once initialized, each candidate $T^{(i)} \in \mathcal{T}$ is refined hierarchically to minimize the alignment error. 
At each pyramid level $\ell$, the transformation is optimized using a robust point-to-plane ICP objective:
\begin{equation}
\min_{T \in \mathrm{SE}(3)} 
\sum_{(\mathbf{m}, \mathbf{o}) \in \mathcal{C}_\ell(T)} 
\rho\!\left(\mathbf{n}_o^\top(R\mathbf{m} + \mathbf{t} - \mathbf{o})\right),
\end{equation}
where $\rho(\cdot)$ is the Huber loss, and $\mathcal{C}_\ell(T)$ denotes correspondences established within a dynamic distance gate $\tau$ and filtered by normal consistency.

\textbf{Scoring and selection.}
Refined candidates are ranked by a quality score $s(T)$ combining spatial coverage and residual error:
\begin{equation}
s(T) = \lambda_c \cdot \mathrm{cov}_\tau(T) 
      + \lambda_r \cdot \left(1 - \frac{\overline{d}_\alpha(T)}{d_{\max}}\right),
\end{equation}
where $\mathrm{cov}_\tau$ is the inlier coverage ratio, $\overline{d}_\alpha$ is the trimmed mean distance of correspondences, and $\lambda_{c,r}$ are weighting factors. 
The candidate with the highest score is selected as the current frame’s registration result.

\textbf{Temporal filtering.}
To ensure AR stability, the selected pose undergoes temporal filtering before visualization. 
Updates are accepted only if they satisfy coverage and stability constraints relative to the previous frame. 
To prevent oscillation between symmetric solutions, a new pose must improve the score by a margin $\Delta s_{\min}$. 
An escape mechanism relaxes these constraints during prolonged inactivity, allowing the system to recover from tracking failures or abrupt motion. 
The resulting temporally filtered pose provides a stable and accurate spatial anchor for \textit{in-situ} AR visualization, linking the preoperative model to the real-world coordinate system.

\section{Experiments}\label{sec:experiments}
\subsection{Experimental Setup}

We evaluated the system on a full-scale 3D-printed human phantom reconstructed from the CVH dataset~\cite{zhang2004chinese}. Three representative robot-assisted procedures were selected for evaluation, including nephroureterectomy, hysterectomy, and cystectomy. Clinical port templates for these procedures were adopted from established surgical guidelines~\cite{Sakhel10, pathak2017comprehensive}.
For system implementation, raw sensor data from the HL2 were streamed using the \texttt{hl2ss} library~\cite{dibene2022hololens} to a remote workstation equipped with a single Nvidia RTX 3090 GPU, a 3.60 GHz Intel\textsuperscript{®} Xeon\textsuperscript{®} W-2223 CPU, and 32 GB of RAM. The system hyperparameters include the PV image resolution (640$\times$360~px at 30~fps) and the depth map resolution (320$\times$288~px at an average of 5~fps). For human surface segmentation, the SAM inference was triggered at 0.2-second intervals. The corresponding segmentation parameters, including the minimum overlap threshold $\tau_{\mathrm{IoU}}$ and pruning sensitivity threshold $\rho_{\max}$, were set to 0.08 and 0.35, respectively. Registration refinement was performed using a multi-resolution pyramid with 3 levels to enable coarse-to-fine alignment. The weighting factors $\lambda_c$ and $\lambda_r$ used for scoring and selecting registration candidates were set to 0.7 and 0.3, respectively, balancing spatial coverage and residual error.
In the following, we first evaluate the performance of real-world 3D reconstruction and then assess the accuracy of CT-to-scene registration.

\subsection{Evaluation of Real-world 3D Reconstruction  and Mapping}\label{eva_reconstruction}

\begin{figure*}[th]
    \centering
    \includegraphics[width=1.0\textwidth]{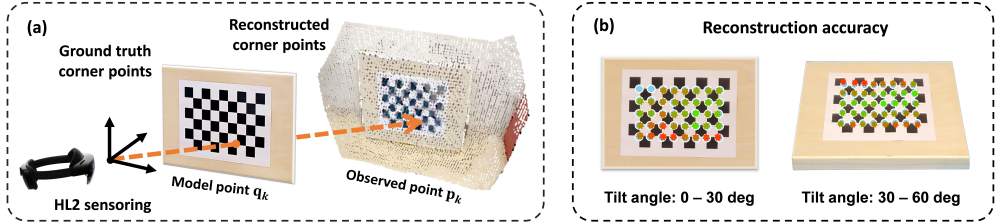}
    \caption{Evaluation of reconstruction. (a) Assessment procedure: ground-truth points \(\mathbf{q}_k\) are directly extracted from the checkerboard; reconstructed points \(\mathbf{p}_k\) are computed from HL2 sensor data. (b) Visualization of reconstruction results: blue points denote the ground truth; colored points encode reconstructed positions and errors.}
    \label{fig:eval_reconstruction}
\end{figure*}

\noindent To quantitatively assess the geometric accuracy of the real-world 3D reconstruction, we use a planar checkerboard with known metric geometry as a standardized ground-truth reference. For each frame, reconstructed checkerboard corners obtained from the HL2 mapping pipeline are compared against the ideal checkerboard model, and the reconstruction error is analyzed with respect to viewing distance and tilt angle (Fig.~\ref{fig:eval_reconstruction}).

At time $t$, the HL2 provides a PV image $I_t$, an LT depth map $Z_t$, and the synchronized reference-to-world pose $T_{\mathrm{ref}\to w}(t)$. Using the mapping procedure described in Sec.~\ref{Real-world 3D Mapping}, the LT depth stream is fused into a colored point cloud in the world frame.

To establish geometric ground truth, we use a checkerboard with $6\times9$ squares, defining an internal grid of $c\times r=8\times5$ corners. The ideal 3D corner locations are defined in the checkerboard coordinate frame as
$\mathbf{q}_{ij} = [\, i d,\; j d,\; 0 \,]^\top$,
with $i=0,\ldots,c{-}1$, $j=0,\ldots,r{-}1$, and grid spacing $d=30$ mm.
For convenience, a linear index $k=i+jc$ is used and we denote $\mathbf{q}_k \triangleq \mathbf{q}_{ij}$.

Checkerboard corners $\{\mathbf{u}_k=(u_k,v_k)\}$ are detected in the PV image with sub-pixel refinement. For each corner, the depth value $z_k$ is estimated as the 30th percentile of nearby samples projected from the LT point cloud to improve robustness. The corresponding 3D point $\mathbf{p}_k$ is obtained by back-projecting $\mathbf{u}_k$ through the PV intrinsics and transforming it into the world frame using the synchronized pose $T_{\mathrm{ref}\to w}(t)$.

To recover the viewing geometry, a Perspective-$n$-Point (PnP) solution is computed from $\{\mathbf{q}_k\}$ and $\{\mathbf{u}_k\}$ to estimate the checkerboard normal. This step is used only to derive the viewing distance and tilt angle, and not as a reference for accuracy evaluation. Reconstruction accuracy is evaluated by estimating the best-fit rigid transform $(R^\ast,\mathbf{t}^\ast)$ aligning the checkerboard model points $\{\mathbf{q}_k\}$ to the reconstructed 3D points $\{\mathbf{p}_k\}$. The per-corner Euclidean residual is then computed as
\begin{equation}
e_k = \bigl\| \mathbf{p}_k - (R^\ast \mathbf{q}_k + \mathbf{t}^\ast) \bigr\|_2 .
\end{equation}
Summary statistics, including mean, median, standard deviation, and extrema, are reported over all samples.

To analyze viewpoint dependency, results are stratified by viewing distance and tilt angle. Viewing distance is defined as the mean depth of the reconstructed checkerboard corners in the PV camera frame. The tilt angle is computed as $\theta=\arccos(n_z)\cdot180/\pi$, where $n_z$ is the $z$-component of the checkerboard normal estimated via PnP. Based on the observed distributions, viewing distances are grouped into Close (0.3--1.0\,m), Medium (1.0--1.5\,m), and Far ($>$1.5\,m), and tilt angles into Low (0--30$^\circ$), Medium (30--60$^\circ$), and High ($>$60$^\circ$). To account for distance-dependent scaling, we additionally report a normalized reconstruction error defined as $\mathrm{NME} = (\mathrm{RMS}_{\mathrm{mm}}/1000)/\overline{Z}$.

\newcommand{\wtext}{15mm} 
\newcommand{\wnum}{12mm}  

\begin{table}[th]
\centering
\caption{Reconstruction accuracy across viewing distances and tilt angles.}
\label{tab:eval-stats}
\footnotesize
\setlength{\tabcolsep}{2.5pt}
\renewcommand{\arraystretch}{1.12}
\begin{tabular}{
  @{} >{\raggedright\arraybackslash}p{\wtext}
      !{\hspace{3pt}\color{gray!60}\vrule width 0.3pt\hspace{3pt}} 
      >{\raggedright\arraybackslash}p{\wtext}
      !{\hspace{3pt}\color{gray!60}\vrule width 0.3pt\hspace{3pt}} 
      *{5}{>{\centering\arraybackslash}p{\wnum}}
      !{\hspace{3pt}\color{gray!60}\vrule width 0.3pt\hspace{3pt}} 
      >{\centering\arraybackslash}p{\wnum}
      @{}}
\toprule
\multirow{2}{*}{\textbf{Distance}} & \multirow{2}{*}{\textbf{Angle}} &
\multicolumn{5}{c!{\hspace{3pt}\color{gray!60}\vrule width 0.3pt\hspace{3pt}}}{\textbf{3D Euclidean Error (mm)}} &
\multirow{2}{*}{\textbf{NME}} \\
\cmidrule(lr){3-7}
& & \textbf{Mean} & \textbf{Std} & \textbf{Median} & \textbf{Min} & \textbf{Max} & \\
\midrule
       & Low     & \good{4.33} & \good{1.00} & 4.67 & \good{2.27} & \good{5.62} & \good{0.0050} \\
Close   & Medium  & 4.64        & 1.10 & \good{4.14} & 3.59 & 7.34 & 0.0055 \\
       & High    & 5.24        & 1.63 & 5.08 & 3.33 & 8.32 & 0.0062 \\
\midrule
       & Low     & 6.90 & 1.86 & 7.57 & 2.92 & 8.68 & 0.0058 \\
Medium & Medium  & 7.22        & 3.67 & 5.82 & 4.40 & 14.74 & 0.0054 \\
       & High    & 13.51        & 1.73 & 14.14 & 9.22 & 15.12 & 0.0094 \\
\midrule
       & Low     & 20.27        & 2.52 & 19.82 & 17.01 & 23.28 & 0.0118 \\
Far    & Medium  & \bad{20.99}        & 1.11 & \bad{20.93} & \bad{19.14} & 22.75 & \bad{0.0121} \\
       & High    & 17.87  & \bad{3.63} & 16.60 & 13.56 & \bad{25.31} & 0.0104 \\
\bottomrule
\end{tabular}
\end{table}

The quantitative results are summarized in Table~\ref{tab:eval-stats}. The system achieves sub-centimeter reconstruction accuracy at close range with low tilt angles, while errors increase with larger viewing distances and tilt angles, primarily due to reduced depth reliability and increased perspective distortion.

Beyond geometric accuracy, we evaluate the system-level performance of the proposed real-world mapping pipeline. Specifically, we quantify the end-to-end latency, spanning from sensor acquisition on the HL2, through remote 3D reconstruction, to local rendering of the scene on the headset. The current implementation achieves an average end-to-end latency of 890\,ms and an update rate of 5\,fps. This measurement encompasses all overheads, including data capture, network transmission, and remote processing, thereby reflecting the real-time efficiency of the complete online pipeline.

\subsection{Evaluation of CT-to-Scene Registration}

We evaluate the accuracy of CT-to-scene registration using a full-scale human phantom fabricated from CT data. A set of 13 anatomical landmarks distributed across different body regions is identified on the CT-derived surface model and used as fiducials for evaluation, including the bilateral eye corners, bilateral mouth corners, bilateral nipples, the navel, bilateral groin landmarks, bilateral fingertips, and bilateral toe tips. These landmarks are geometrically salient on the human body surface and can be consistently localized on both the CT model and the physical phantom. Their 3D coordinates in the CT model are denoted as $\{\mathbf{x}_i^{\mathrm{CT}}\}_{i=1}^{N}$. The corresponding fiducials on the physical phantom are digitized using a tracked stylus with sub-millimeter accuracy, yielding their 3D positions $\{\mathbf{x}_i^{\mathrm{W}}\}_{i=1}^{N}$ in the world coordinate frame.

\begin{figure*}[h]
    \centering
    \includegraphics[width=1.0\textwidth]{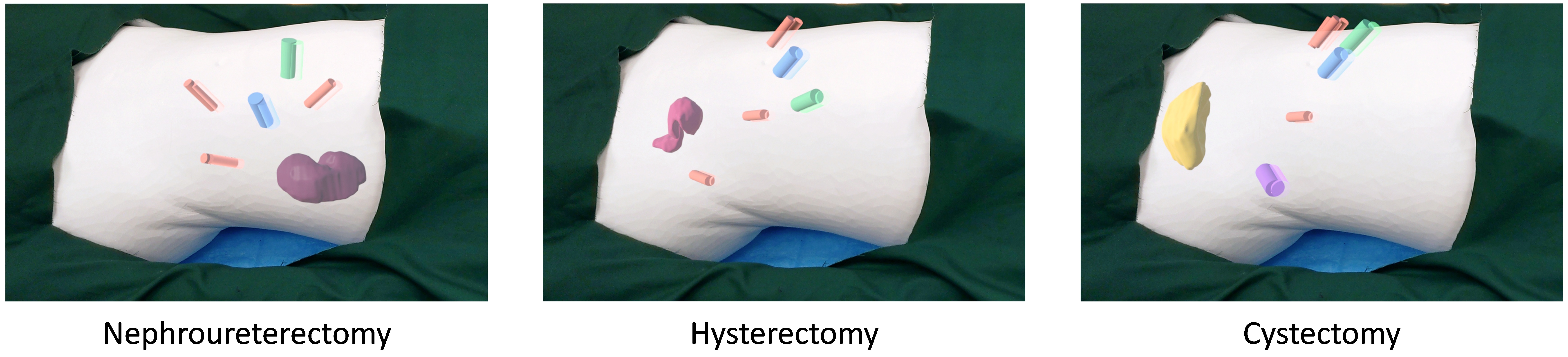}
    \caption{Visualization of cannula alignment on the phantom. Cannulas transformed using the fiducial-based reference alignment are shown as semi-transparent, while those obtained from the markerless AR-based registration are shown as solid.}
    \label{fig:eval_registration}
\end{figure*}

A fiducial-based reference alignment $T_{\mathrm{fid}} \in SE(3)$ is obtained by aligning the CT fiducial positions ${\mathbf{x}_i^{\mathrm{CT}}}$ with their digitized counterparts ${\mathbf{x}_i^{\mathrm{W}}}$ using a least-squares formulation, serving as a reference derived from physical ground-truth measurements. Independently, the proposed ARport pipeline estimates a rigid transformation
$T_{\mathrm{est}} \in SE(3)$
based solely on surface reconstruction and geometric registration, without using any fiducial or marker information. Registration accuracy is quantified using the target registration error (TRE) evaluated at the fiducial locations:
\begin{equation}
\mathrm{TRE}_i =
\left\|\,
\bigl( R_{\mathrm{est}}\,\mathbf{x}_i^{\mathrm{CT}} + \mathbf{t}_{\mathrm{est}} \bigr)
- \mathbf{x}_i^{\mathrm{W}}
\,\right\|_2 .
\end{equation}
The TRE measures the Euclidean distance between the fiducial location predicted by the markerless AR-based registration and the corresponding digitized location on the physical phantom. To characterize the spatial variation of registration accuracy, we additionally compute two geometric descriptors for each landmark. The first is the distance to the visible area (DVA), defined as the Euclidean distance from the landmark to the nearest surface region observed by the HL2. The second is the distance to the torso mid-sagittal plane (DMP), defined as the perpendicular Euclidean distance from the landmark to the plane that bisects the torso into left and right halves.

Table~\ref{tab:ct-registration-stats} reports summary statistics of the TRE together with the corresponding DVA and DMP values. Overall, landmarks closer to the visible region and nearer to the torso mid-sagittal plane tend to exhibit smaller TRE values, while larger errors are observed for landmarks located farther away. This behavior is consistent with the fact that the surface observed during registration is primarily distributed over a cylindrical-like abdominal region, resulting in reduced sensitivity to small rotations about the cranio--caudal axis. For intuitive visualization, Fig.~\ref{fig:eval_registration} shows the overlaid pre-planned cannula layouts on the physical phantom, transformed using the fiducial-based reference alignment $T_{\mathrm{fid}}$ and the markerless AR-based estimate $T_{\mathrm{est}}$.

\newcommand{\wtexts}{12mm}
\newcommand{\wtextl}{18mm}
\newcommand{\wnumr}{10mm}  
\begin{table}[t]
\centering
\caption{Registration accuracy across body regions and anatomical landmarks.}
\label{tab:ct-registration-stats}
\footnotesize
\setlength{\tabcolsep}{2.5pt}
\renewcommand{\arraystretch}{1.12}
\begin{tabular}{
  @{} >{\raggedright\arraybackslash}p{\wtexts}
      !{\hspace{3pt}\color{gray!60}\vrule width 0.3pt\hspace{3pt}}
      >{\raggedright\arraybackslash}p{\wtextl}
      !{\hspace{3pt}\color{gray!60}\vrule width 0.3pt\hspace{3pt}}
      *{5}{>{\centering\arraybackslash}p{\wnumr}}
      !{\hspace{3pt}\color{gray!60}\vrule width 0.3pt\hspace{3pt}}
      >{\centering\arraybackslash}p{\wnumr}   
      !{\hspace{3pt}\color{gray!60}\vrule width 0.3pt\hspace{3pt}}
      >{\centering\arraybackslash}p{\wnumr}   
      @{}}
\toprule
\multirow{2}{*}{\textbf{Region}} & \multirow{2}{*}{\textbf{Landmark}} &
\multicolumn{5}{c!{\hspace{3pt}\color{gray!60}\vrule width 0.3pt\hspace{3pt}}}{\textbf{Target Registration Error (mm)}} &
\multirow{2}{*}{\makecell{\textbf{DVA}\\\textbf{(cm)}}} &
\multirow{2}{*}{\makecell{\textbf{DMP}\\\textbf{(cm)}}} \\
\cmidrule(lr){3-7}
& & \textbf{Mean} & \textbf{Std} & \textbf{Median} & \textbf{Min} & \textbf{Max} & & \\
\midrule

\multirow{2}{*}{Head}
 & Eye corners  & 6.96 & 2.67 & 6.66 & 3.16 & 12.16 & 38.52 & 1.76 \\
 & Mouth corners & 6.41 & 2.77 & 7.23 & 3.36 & 8.99  & 32.67 & 2.21 \\
\midrule

\multirow{3}{*}{Torso}
 & Nipples & 8.05 & 2.36 & 8.13 & 3.35 & 11.81 & 9.18  & 10.03 \\
 & Navel  & \good{5.98} & \good{1.37} & \good{5.99} & 4.33 & \good{8.40} & \good{0.12} & \good{0.22} \\
 & Groin  & 7.82 & 3.05 & 8.35 & 3.04 & 12.37 & 4.87  & 5.70 \\
\midrule

\multirow{2}{*}{Limbs}
 & Fingertips & 8.34 & 4.22 & 8.45 & \good{2.66} & 14.00 & 4.67  & \bad{17.46} \\
 & Toe tips & \bad{17.81} & \bad{4.36} & \bad{16.95} & \bad{10.12} & \bad{24.43} & \bad{83.35} & 14.57 \\
\bottomrule
\end{tabular}
\end{table}

\section{Discussion and Conclusion} 
This study addresses a practical need in robot-assisted surgery: translating preoperative port placement plans onto the patient’s body surface during surgical setup. We present ARport, an image-guided system that visualizes planned trocar sites \textit{in-situ} through augmented reality.

\textbf{Proof of concepts}. ARport implements a complete end-to-end pipeline that acquires raw sensor data from the HL2, performs remote computation for scene reconstruction, foundation model-based segmentation, and markerless CT-to-scene registration, and ultimately renders planned port sites directly on the patient’s body surface. These components are integrated into a coherent and efficient workflow that operates without additional tracking hardware or manual alignment, enabling seamless integration into the surgical setup process.
Experiments conducted on a full-scale human phantom demonstrate accurate and stable port placement under rigid conditions, with sub-centimeter localization error in the clinically relevant torso region where trocar insertion is performed. This level of accuracy is particularly relevant compared to conventional heuristic-based port placement in current clinical practice, which commonly relies on manual estimation methods such as the finger-breadths rule ~\cite{harper2011standardized} and is known to introduce centimeter-level variability across operators~\cite{rincon2020estimation}. By reducing this initial placement uncertainty, ARport provides intuitive \textit{in-situ} localization, improving the accuracy and reproducibility of port placement during surgical setup. Together, these results validate the technical feasibility of ARport and highlight its potential to enhance spatial guidance in robot-assisted surgery.

\textbf{Limitations and future works}. While the proposed system demonstrates feasibility in a controlled phantom setting, several limitations remain to be addressed before clinical deployment. First, the current registration error of ARport (ranging from 5--10~mm) remains higher than that of conventional surgical navigation systems, which typically maintain localization errors within approximately 5~mm \cite{han2024review}. This accuracy gap is primarily attributed to the systematic measurement bias of the HL2 depth sensor identified in our reconstruction evaluation (Sec.~\ref{eva_reconstruction}), which varies with sensing distance and angle. Since prior studies \cite{li2024evd,yang2025easyreg} have successfully mitigated such raw depth errors to sub-millimeter levels through patient-specific calibration, addressing this hardware-induced bias offers a promising path to improve ARport’s registration performance. Second, the current evaluation was conducted under rigid phantom conditions and does not account for the soft-tissue compliance of the abdominal wall. In clinical practice, the introduction of pneumoperitoneum routinely causes non-rigid deformation, potentially altering the spatial correspondence between preoperative plans and intraoperative physical sites \cite{maddah2020measuring,yang2025learning, yang2024boundary,han2025towards}. Future work will involve developing deformation compensation algorithms and validating them on a realistic and deformable phantom that simulates the pneumoperitoneum.
Third, the robustness of ARport under real-world operating room conditions requires further investigation. Factors such as dynamic lighting, surgical drapes, and instrument occlusion may potentially compromise the feasibility of SAM-based segmentation and the robustness of FPFH descriptors during the registration process \cite{guo2016comprehensive}. To improve clinical adaptability, we plan to collect clinical datasets to validate and refine each module for realistic environments. In addition, a systematic evaluation of pose stability over time will be conducted to ensure that the registration results remain consistent during prolonged use, especially under dynamic motion or subtle changes in the environment.
Fourth, as ARport is deployed on an OST-HMD, its clinical utility is heavily dependent on usability and human factors, which were not assessed in this preliminary study. Future work will include formal user studies to evaluate the system’s impact on surgical workflow, cognitive load, and ergonomic comfort. Such assessments are critical to ensuring that the AR interface provides intuitive guidance without introducing visual fatigue or distraction, ultimately facilitating a seamless transition from laboratory prototypes to practical clinical tools.

\textbf{Conclusion}. ARport provides an AI-enabled AR system for markerless, image-guided port placement in robotic surgery. It demonstrates feasibility through a comprehensive phantom study. Future work will focus on improving registration accuracy by mitigating sensor-induced reconstruction bias in OST-HMDs, incorporating deformation compensation to account for pneumoperitoneum, validating the system on patient data, and conducting formal user studies to assess usability, workflow integration, and ergonomics, toward translating ARport into a practical clinical tool.

\section*{Acknowledgements}
This study was supported in part by Hong Kong Innovation and Technology Fund Project No. GHP/167/22SZ, and in part by the InnoHK initiative of the Innovation and Technology Commission of the Hong Kong Special Administrative Region Government.

\bibliography{sn-bibliography}

\end{document}